\documentclass{article}

     \PassOptionsToPackage{compress}{natbib}
    \usepackage[final, nonatbib]{neurips_2020}

\usepackage[utf8]{inputenc} 
\usepackage[T1]{fontenc}    
\usepackage{xurl}            
\usepackage{booktabs}       
\usepackage{amsfonts}       
\usepackage{nicefrac}       
\usepackage{microtype}      
\usepackage{xcolor}
\usepackage[colorlinks = true, 
            linkcolor = blue,
            urlcolor  = teal,
            citecolor = blue]{hyperref}
\usepackage{graphicx}
\usepackage{caption}
\usepackage{subcaption}
\usepackage{grffile}
\usepackage{amsmath}

\newcommand{\cmmnt}[1]{\ignorespaces}

\title{Understanding Learned Reward Functions}

\author{%
  Eric J. Michaud\\
  UC Berkeley \\
  \texttt{eric.michaud99@gmail.com} \\
  \And
  Adam Gleave \\
  UC Berkeley \\
  \texttt{gleave@berkeley.edu} \\
  \And
  Stuart Russell \\
  UC Berkeley \\
  \texttt{russell@berkeley.edu} \\
}

\begin{document}

\maketitle

\begin{abstract}
In many real-world tasks, it is not possible to procedurally specify an RL agent's reward function. In such cases, a reward function must instead be \emph{learned} from interacting with and observing humans. However, current techniques for reward learning may fail to produce reward functions which accurately reflect user preferences. Absent significant advances in reward learning, it is thus important to be able to audit learned reward functions to verify whether they truly capture user preferences. In this paper, we investigate techniques for \textit{interpreting} learned reward functions. In particular, we apply saliency methods to identify failure modes and predict the robustness of reward functions. We find that learned reward functions often implement surprising algorithms that rely on contingent aspects of the environment.
We also discover that existing interpretability techniques often attend to irrelevant changes in reward output, suggesting that reward interpretability may need significantly different methods from policy interpretability.
\end{abstract}

\section{Introduction}

\subsection{Reward Learning}
\label{sec:reward_learning_intro}

In the last several years, reinforcement learning techniques have produced agents surpassing human performance in tasks as varied as Atari games~\cite{mnih2015human}, DOTA~\cite{OpenAI_dota}, Go \cite{silver2016mastering} and Starcraft~\cite{alphastar}. In each of these tasks, it is possible for a designer to write down an appropriate reward function. However, in complex real-world tasks, it is often difficult to manually and accurately translate human preferences and goals into a reward function. Attempts at manually specifying an agent's objective can fail in a variety of ways. For instance, the stated objective could express indifference over environmental variables actually relevant to users, or the reward function could contain exploitable quirks leading to unexpected and harmful agent behavior that nevertheless maximize the reward~\cite{amodei2016concrete}.

In such cases, human preferences should instead be \textit{learned} by agents via a process of interacting with and observing humans. This task has been formalized by frameworks such as assistance games and POMDPs~\cite{fickinger2020multiprincipal, hadfield2016cooperative, anonymous2021benefits} and reward-rational choice \cite{jeon2020rewardrational}. Inverse Reinforcement Learning (IRL) infers a reward function $R$ for which observed user demonstrations are optimal. Other reward learning techniques learn a reward function not by observing user behavior, but by soliciting user preferences between agent trajectories~\cite{christiano2017deep}. Some algorithms use both user demonstrations and preference comparisons to train a reward model~\cite{ibarz2018reward, brown2019extrapolating}.

Despite much progress, reward learning algorithms still often fail to produce satisfactory reward functions. An important failure mode of reward learning, and of machine learning in general, occurs when models depend on spurious features. Learned reward functions could, for instance, detect features of the environment which are correlated with the return of reward but not causally connected with it. Such models are brittle, as the spurious correlations may break down in novel situations.

For an example of this (adapted from \cite{haan2019causal}), consider an agent performing reward learning for the task of driving a car. Observing that whenever a human driver brakes, the brake indicator on the dashboard turns on, the agent learns to assign positive reward to braking when the brake indicator is on. However, the brake indicator turning on, while correlated with reward, is not ultimately \emph{why} the human is braking the car. A policy trained on this reward function would likely be unsafe. 

Examples like these motivate us to find methods of auditing a machine's understanding of our preferences. At the very least, we should be able to verify that the machine is attuned to features of the environment which are truly relevant.

\subsection{On the simplicity of reward functions vs.\ policies}
One reason to analyze learned reward functions, and not just policies trained on them, is that reward functions may be easier to understand than policies. As RL algorithms increase in sophistication and produce increasingly capable agents, these agent's policies could also become increasingly difficult to understand mechanistically \cite{hubinger2019chris}. It may be a property of intelligent systems generally, and highly competent ones especially, that while their objectives may be simple, their means of achieving these objectives could be quite complex. In \textit{Artificial intelligence as a positive and negative factor in global risk} \cite{yudkowsky2008artificial}, Yudkowsky provides the following example:

\begin{quote}
    An optimization process steers the future into particular regions of the possible. I am visiting a distant city, and a local friend volunteers to drive me to the airport. I do not know the neighborhood. When my friend comes to a street intersection, I am at a loss to predict my friend’s turns, either individually or in sequence. Yet I can predict the result of my friend’s unpredictable actions: we will arrive at the airport... I can predict the outcome of a process, without being able to predict any of the intermediate steps in the process.
\end{quote}

In this example, the objective "get to the airport" is simpler to describe than the task of \emph{how} to get to the airport. Across many environments, it may be easier to understand reward functions than policies -- objectives are simpler than the algorithm learned to achieve them. Interpreting policies is still important -- it allows us to detect failure modes introduced by imperfect optimization. However, if reward learning algorithms fail to learn a user's true reward function, it may be easier to identify that failure by directly analyzing the reward function, and not the policy trained on it.

This is also not to say that human objectives will be simple -- indeed, we expect them to be too complicated to manually describe. However, we expect the complexity of human objectives to be constant or grow only slowly, whereas the complexity of agents could grow quickly as they become more capable. Thus we expect the gap between the simplicity of reward functions vs. the complexity of policies to increase over time.

\subsection{Evaluating learned reward functions}
\label{sec:evaluating_functions}

When evaluating reward learning algorithms, it is common to look at rollouts from a policy trained on the learned reward function. However, if the rollouts fail to achieve the desired goal, it is often unclear whether the algorithm has failed to learn an appropriate reward function or whether the RL algorithm just failed to learn a good policy given the learned reward function. It is therefore desirable to instead directly evaluate the learned reward function. 

Gleave et al.~\cite{gleave2020quantifying} propose comparing learned reward functions to a ground-truth reward using a pseudometric called EPIC. This is useful for benchmarking algorithms, where a ground-truth reward is often available.
However, reward learning algorithms are useful precisely when the ground-truth reward function \emph{cannot} be manually specified.
Thus EPIC, on its own, gives little insight into the accuracy of learned reward functions in the real world. 


Ideally, we should like to know whether a learned reward function fails to reflect human preferences before training and certainly before deploying a policy. In this work, we adapt and empirically investigate interpretability techniques for learned reward models. To the best of our knowledge, this is the first application of neural network interpretability methods to reward functions.
Using a combination of saliency techniques and counterfactual inputs, we find that learned reward functions sometimes implement surprising algorithms.
Moreover, we find saliency maps can often identify when reward functions use robust vs.\ spurious features, and furthermore can predict when the function output will change in response to environment modifications. However, these techniques are conservative: some changes they detect may not prevent a policy trained on the learned reward function from being successful. This is related to the well-known result that some transformations, such as potential shaping $R(s, a, s') \mapsto R(s, a, s') + \gamma\Phi(s') - \Phi(s)$, leave the optimal policy of $R$ unchanged~\cite{ng1999policy}. Despite this limitation, we believe our methods are still useful, and that they illustrate the relevance of interpreting learned reward functions and the necessity for future work in this area.

\section{Related Work}
Prior work on interpretability for reinforcement learning has focused on policies, not reward functions. Many techniques have been investigated. Rupprecht et al.  \cite{rupprecht2019finding} train a generative model over the state space of Atari games to visualize states which minimize or maximize given action probabilities. Greydanus et al. \cite{greydanus2018visualizing} analyze Atari policies by blurring out regions of observations and recording how much the value and policy networks outputs change. This produces a heat map over the image of the importance of each region. Mott et al. \cite{mott2019towards} analyze agents by adding an attention bottleneck to the architecture of the policy, allowing the information the agent is using to be more directly visualized. 

The closest work we are aware of is by Russell and Santos \cite{russell2019explaining}. They analyze reward functions by fitting decision trees to a given reward function and then use a combination of Gini importance and LIME \cite{ribeiro2016should} to identify features of "local" and "global" importance (where "local" features are salient to the prediction of reward on an individual state, and "global" features are salient across all states of the environment). They claim that if the ordering of the locally important and globally important features are well correlated, then these features can be trusted as explanations of the reward function. Unlike their method, which fits a simplified model to a reward function, we apply interpretability techniques directly to learned reward functions. We feel that this direct analysis of reward functions is important for identifying safety failures, as the reward function may harbor failure modes that are not captured in a simplified model.

\section{Methods}
\label{sec:methods}

To analyze reward functions, we employed the raw gradient saliency method~\cite{simonyan2013deep} and an occlusion map method~\cite{greydanus2018visualizing}, as well as counterfactual examples~\cite{goyal2019counterfactual, wachter2017counterfactual}.\footnote{All code for our experiments is available at: \url{https://github.com/HumanCompatibleAI/interpreting-rewards}}

\paragraph{Raw Gradient Saliency} For a given a reward function $R$, we compute $\frac{\partial R}{\partial (s, a, s')}$, that is, the gradient of reward w.r.t. each input. This produces a saliency map over the input space of the reward function, for a given transition $s \xrightarrow{a} s'$. While the gradient only gives information about the local sensitivity of the reward function to small changes in the state of the environment, in practice it often produces sensible and interpretable saliency maps that pass basic sanity checks \cite{adebayo2018sanity}.

\paragraph{Occlusion Maps} This method performs a Gaussian blur across different regions of the observations and records how much this perturbation changes the predicted reward, producing a heat map. One way of interpreting these blurring perturbations is that they remove information about where features in the environment are precisely located, or how features change over time (if observations consist of multiple images). The heat maps therefore reflect the sensitivity of the reward function to the loss of such information in each region of the input image. This method is best suited for environments with large image-based observation spaces.

\paragraph{Counterfactual Inputs} We pass hand-crafted inputs, that may not naturally occur, to our reward functions. These examples allow us to test hypotheses about what the reward function is doing, sometimes providing more insight into reward functions than saliency maps alone.

\paragraph{Method Selection} Saliency maps are best suited to hypothesis generation, providing a high-level overview of relevant features. Counterfactual inputs allow one to test hypotheses, obtaining more detailed understanding, but are more time-consuming to construct.

The appropriate saliency map method largely depends on the scale of the features that are likely relevant to a reward function in a given environment. The raw gradient saliency method is effective when the input dimensions are largely independent of each other. By contrast, when working with high-dimensional images, the relevant features of the image often consist of contiguous blocks of pixels, and the reward function may be independent of changes to individual pixels. In such cases, an occlusion map -- with an appropriate scale blur -- may be easier to interpret. We investigated a wider range of techniques, including integrated gradients \cite{sundararajan2017axiomatic}, Guided Grad-CAM \cite{selvaraju2016gradcam}, DeepLift \cite{shrikumar2017learning} and Shapley values \cite{castro2009polynomial}, but found them to perform poorly in our setting.


\begin{figure*}[t]
    \centering
    \includegraphics{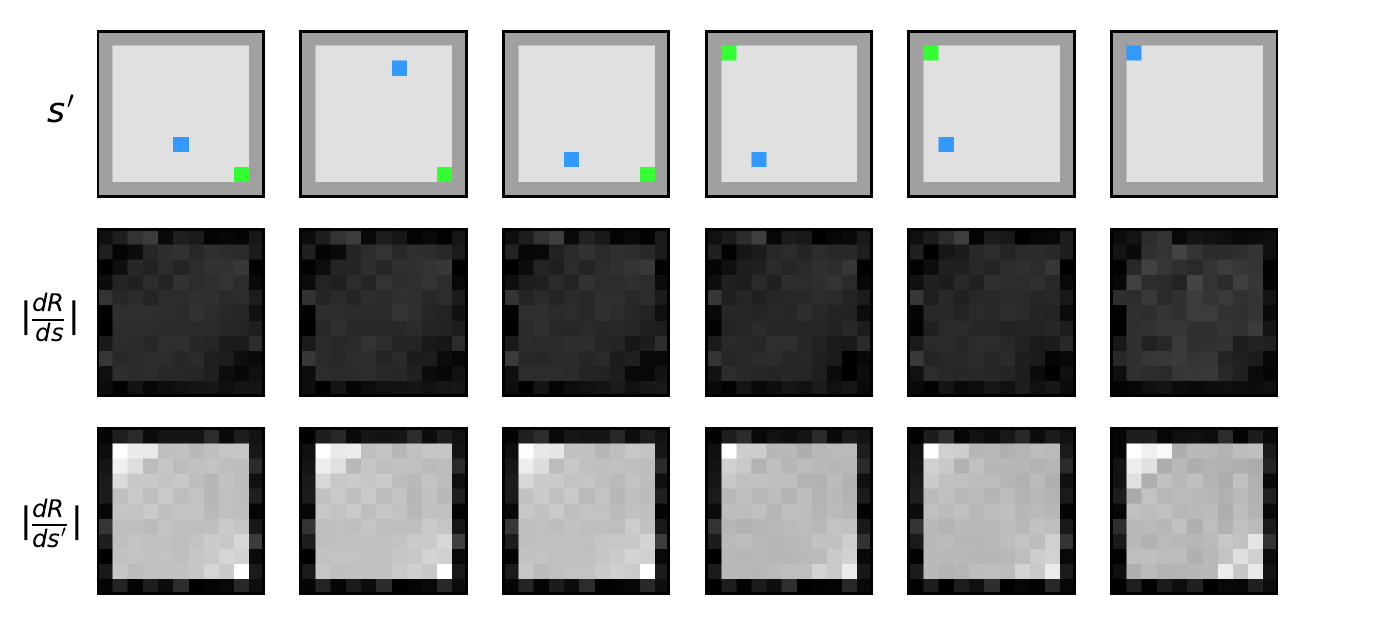}
    \caption{Saliency maps for six transitions in a gridworld environment. In the top row, we show states $s'$ of the environment from transitions $s \rightarrow s'$. For clarity, the agent is shown in blue and the goal is green, however the true observations are greyscale images. In the next two rows, we show the magnitude of the gradient of reward function $R$ with respect to $s$ and $s'$ respectively. Perhaps surprisingly, we observe the learned reward function is sensitive principally to the pixels in $s'$, and not those in $s$. The true reward function for this environment determines the goal position from $s$ and only gives reward if the agent is covering it in $s'$. However, from the saliency map, the learned reward functions seems to mostly attend to information in $s'$, suggesting that it has learned a different algorithm for computing reward.}
    \label{fig:maze_reward_saliency_maps}
\end{figure*}

\section{Experiments}

We analyze reward functions in two image-based environments: simple gridworld environments, and Atari 2600 games. In these environments, we train reward functions via regression on the ground truth reward for the environment on transitions from rollouts of an expert policy. One would expect this method to produce higher-fidelity learned rewards than more complex learning algorithms on noisy and often biased human data. It is therefore striking that, in spite of this, our methods often find that the learned reward implements a surprising algorithm which would fail under distribution shift.

\subsection{Gridworld Environments}

\begin{figure*}[ht]
    \centering
    \includegraphics{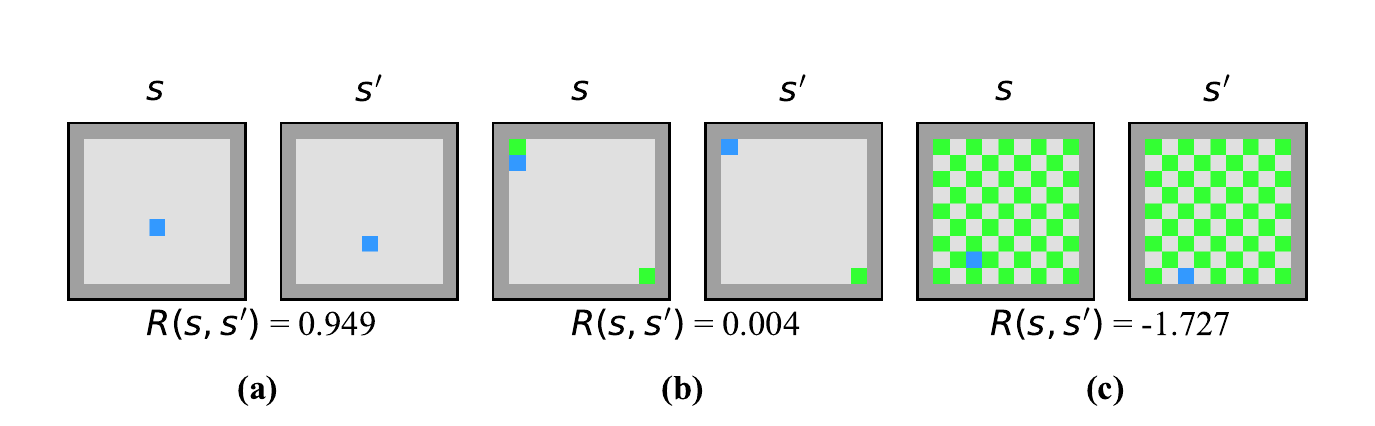}
    \caption{Handcrafted counterfactual transitions for the reward function from Figure \ref{fig:maze_reward_saliency_maps}. In \textbf{(a)}, we remove the goal object from the environment entirely, and the reward function outputs a value close to $1$ when it should be $0$. In \textbf{(b)}, we have two goal objects in the environment, and the reward function returns a value close to $0$ when it should be $1$ (since the agent has just reached one of the two goal positions). In \textbf{(c)}, we add many goal objects to the environment, and the reward function returns a negative value, when it should be $1$. }
        \label{fig:maze_reward_counterfactuals}
\end{figure*}

We start by analyzing simple gridworld environments where an agent attempts to navigate to a goal block. Observations are $11 \times 11$ greyscale images, and the agent receives reward $1$ when it reaches and covers the goal block, and $0$ reward otherwise. Since the relevant features are individual input dimensions, we use raw gradient saliency maps rather than occlusion maps~(see section~\ref{sec:methods}).

We first analyze a version of the environment \texttt{CoinFlipGoal}, where the goal position is chosen randomly between the top left and bottom right of the environment at the beginning of each episode. Our learned reward function takes the form $R(s, s')$, and is trained via regression as previously described. Computing the raw gradient $\frac{\partial R}{\partial (s, s')}$, we find from Figure~\ref{fig:maze_reward_saliency_maps} that our learned reward function appears to be principally sensitive to the pixels in $s'$ and not those in $s$, as the magnitude of the gradient is far greater in $s'$ than $s$.

This result was surprising to us, since it is inconsistent with with how the code for the underlying environment computes reward. The true algorithm for the environment's reward function is:
 \begin{quote}
     \texttt{Are any goal blocks (visible in $s$) covered by the agent in $s'$? If so, output 1, otherwise output 0.}
 \end{quote}
However, since our learned reward function seems to be only looking at pixels in $s'$, it can't be implementing this algorithm. Based on the saliency map, we conjectured that our reward function had instead learned: 
\begin{quote}
    \texttt{If a goal block is visible anywhere in $s'$, output 0, otherwise output~1.}
\end{quote}

To verify that this is what our reward function is doing, in Figure~\ref{fig:maze_reward_counterfactuals} we hand-crafted counterfactual transitions $(s, s')$ and fed them to the reward function. We found that removing the goal block from the environment entirely made the reward function output $1$ (as if the agent had just reached the goal position). Adding a second goal block to the environment (so that goals are in both the top left and bottom right), made the reward function output $0$, despite the agent reaching one of the goal blocks in $s'$. Adding many goal blocks made the reward function output a large negative number. 

These results show the learned reward function is significantly different from the underlying environment's true reward. Instead of returning positive reward in the absence of a visible goal block, it returns negative reward in the presence of goal blocks. This example provides a proof-of-concept that saliency maps can identify relevant features, and that counterfactuals can confirm the effect these features have on the reward output. 

\begin{figure*}
    \centering
    \includegraphics{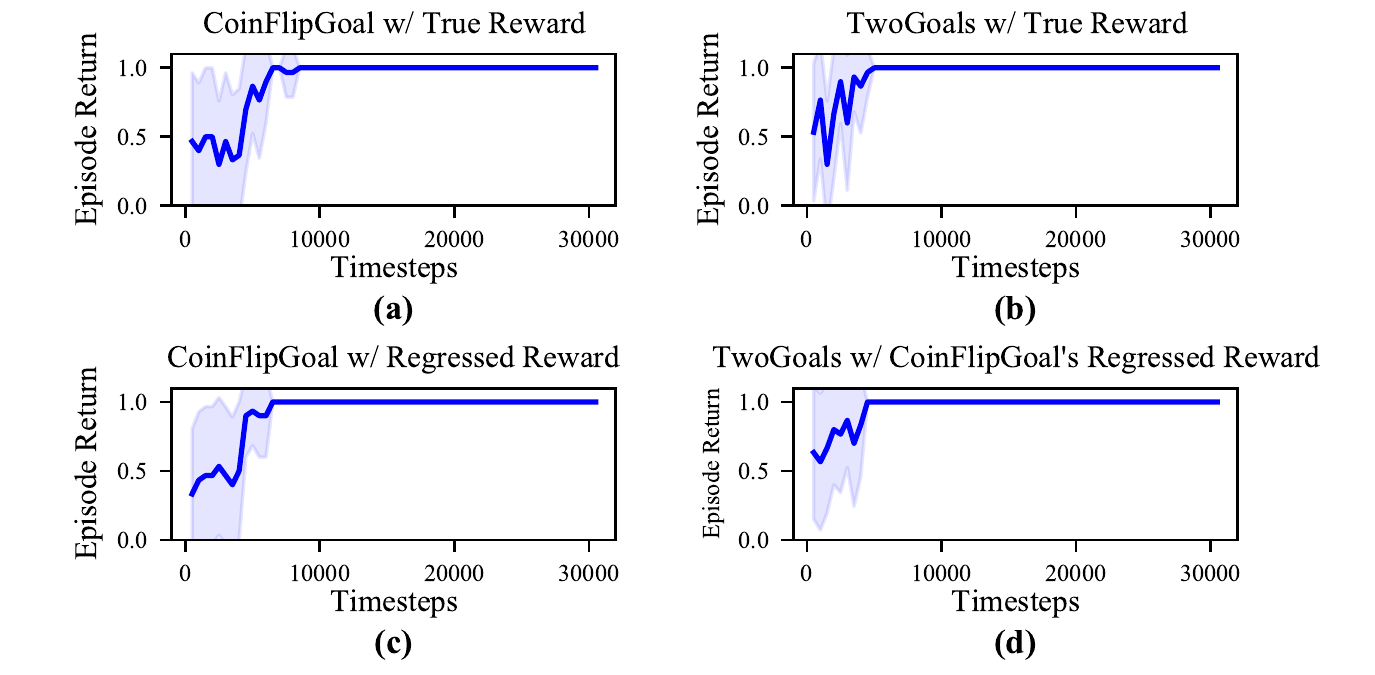}
    \caption{\textbf{Training curves for PPO gridworld agents.} The displayed "Episode Return" is on the ground-truth reward. Despite the addition of a second goal block distorting the outputs of a regressed reward model in an off-distribution environment, the reward model still produces a good policy! Confidence intervals represent variance in the estimate in this policy return, due to a finite number of evaluation episodes. }
    \label{fig:maze_training_curves}
\end{figure*}

Despite implementing a different function, on transitions from \texttt{CoinFlipGoal} the learned reward closely matches the output of the true reward.
However, reward functions that attend to spurious correlations can be fragile.
To investigate this further, we create a new environment, \texttt{TwoGoals}, with goals in both the top left and bottom right corners. The underlying reward structure for this environment is the same as for \texttt{CoinFlipGoal} ($1$ when the agent reaches a goal, and $0$ otherwise).
However, our learned reward function, trained on transitions from \texttt{CoinFlipGoal}, \textbf{outputs significantly different values}. When the agent reaches a goal block, it outputs values $\approx 0$. When it is not on a goal block, it outputs slightly negative values. Despite these substantial differences in output, Figure~\ref{fig:maze_training_curves} show that policies trained on the true and learned reward in have almost identical returns in both \texttt{CoinFlipGoal} and \texttt{TwoGoals}.

This example illustrates that, in any given environment, many different reward functions may incentivize the same behavior.
Existing interpretability techniques alone can identify when and how a reward function's output changes, but cannot directly predict what effect -- if any -- this has on policy behavior.
In some sense, learned reward functions can sometimes get lucky: they can implement the wrong algorithm, have their outputs totally change with small changes in the environment, and yet still lead to similar policies to the true reward in the modified environment.

However, it is possible to envision changes to the environment where this reward function would fail to transfer. For instance, imagine that there was an object in the environment which, once reached by the agent, removed all the goal blocks from the environment. Our learned reward function would incentivize the agent to go to this goal-destroying object. But this behavior would achieve $0$ return on the environment's true reward, and thus our learned reward function would fail catastrophically in this new environment.

%

\subsection{Atari 2600}


We now turn our attention to investigating reward functions in Atari games. In the  Atari environments presented here, $1$ reward is returned when a point is scored and $0$ otherwise. We use the Nature DQN preprocessing \cite{mnih2015human}: converting the image observations into $84 \times 84$ greyscale and stacking the four latest frames on top of each other, forming an observation space of shape $4 \times 84 \times 84$. Our reward functions are CNNs, take the form $R(s')$, and are trained via regression on synthetic ground-truth reward labels. We use the Gaussian-blur occlusion technique of \cite{greydanus2018visualizing} (with $\sigma = 3$) since relevant features in Atari tend to consist of large contiguous blocks (see~\ref{sec:methods}).

\begin{figure}[ht]
    \includegraphics{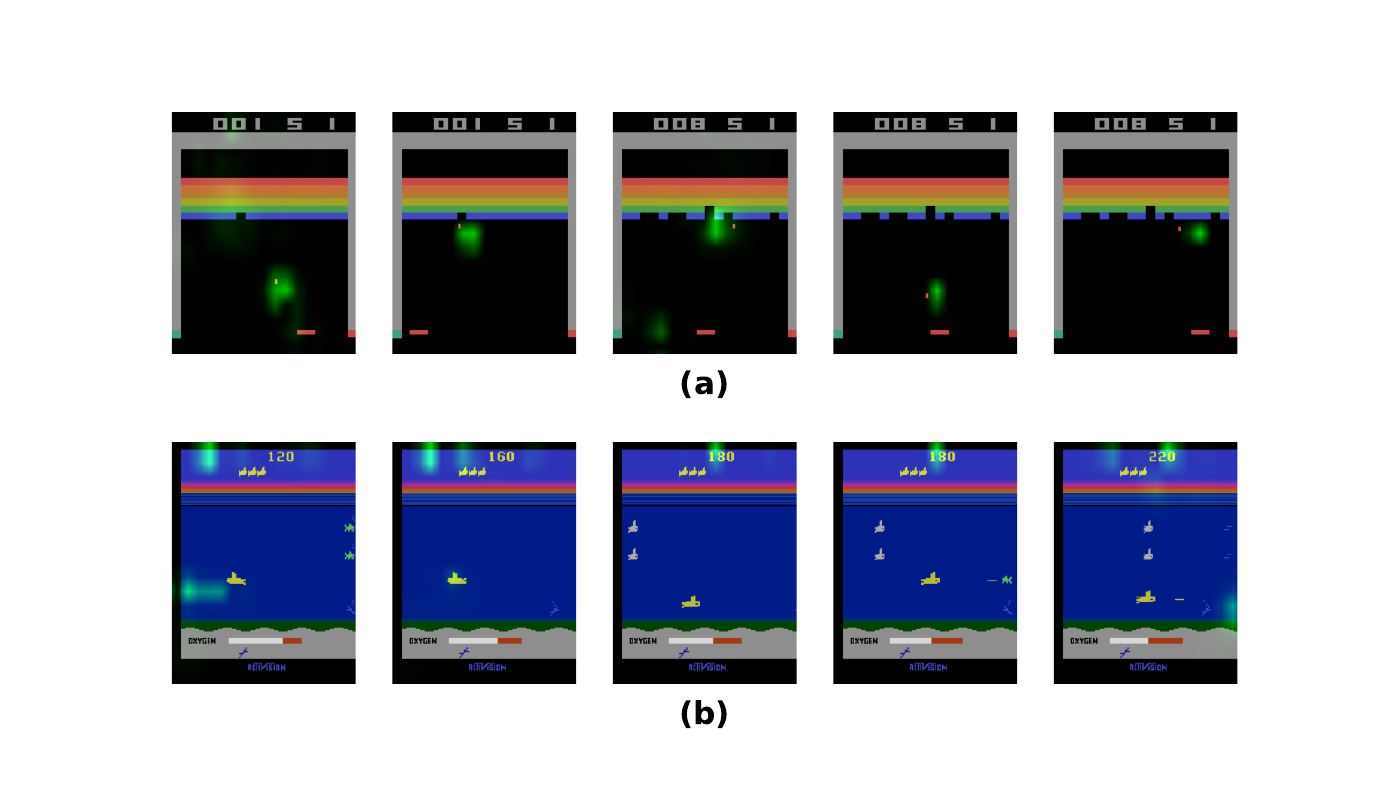}
    \caption{\textbf{Occlusion maps in Atari 2600}. Overlaid in green, the heat map from the Gaussian-blur perturbations shows that the ball position is most consistently salient to our Breakout reward model \textbf{(a)}, and that the score display is most salient to our Seaquest reward model \textbf{(b)}.}
    \label{fig:atari-saliency-maps}
\end{figure}

In Figure~\ref{fig:atari-saliency-maps}(a) we find that in Breakout the ball position tends to be most salient to the learned reward function. This makes sense:  if the ball is detected away from the blocks, reward $0$ can be confidently returned, and reward $1$ should be returned when the ball is both close to the blocks and moving toward them. However, we were surprised to see from Figure~\ref{fig:atari-saliency-maps}(b) that in Seaquest the region most consistently salient to the reward function was the game's score display.

Breakout also contains a score, so one might wonder why in this environment the learned reward attends to the ball and not the score.
It turns out that the reason for this discrepancy is that in Breakout, the score displayed doesn't update until the time step \emph{following} the return of reward.
By contrast, in Seaquest the displayed score updates immediately in $s'$.
Thus the Seaquest reward function can simply see whether the score pixels \emph{change} to predict reward -- it need not even understand the digits.

This result suggests that when testing reward learning algorithms in Atari environments, the score display should be removed from the environment. If the score is displayed, the task of predicting reward may not require that the model learns anything complex about the structure of the environment -- it can just look at the score. In real life, the task of predicting reward is more complicated than this. 

\begin{figure*}[ht]
    \includegraphics{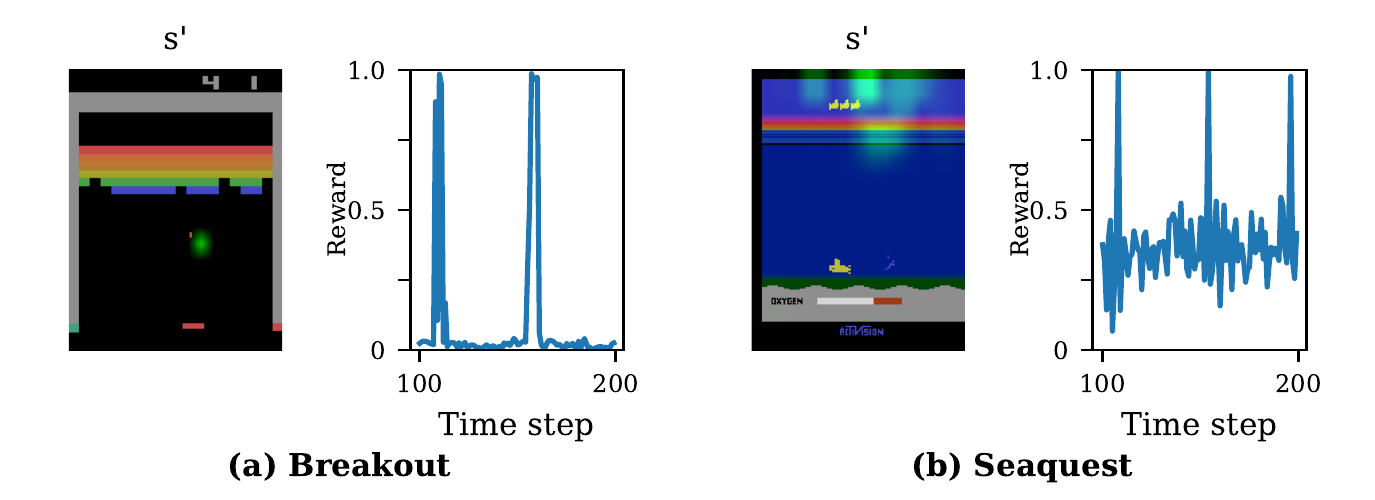}
    \caption{\textbf{Time series of predicted reward}, for reward models trained with the score displayed, but evaluated on environments with the score removed. Removing the score from Breakout doesn't effect the reward function output (it still correctly outputs 0 or 1 at the appropriate times), but for Seaquest removing the score significantly changes the function's output -- it is far noisier after removing the score.}
    \label{fig:atari_timeseries}
\end{figure*}

Indeed, when we remove the score from the environment, but use reward models trained on observations with the score present, we see from Figure~\ref{fig:atari_timeseries} that outputs for the Breakout reward model are unchanged, whereas the Seaquest reward model becomes quite noisy. However, as with our gridworld environments, although saliency predicts the sensitivity of our reward functions to perturbations of the environment, it does not illuminate whether the reward function will transfer to the modified environment in the relevant sense: whether its optimal policy will coincide with the true optimal policy in the modified environment. Reward functions can output totally different values when transferred to a slightly different environment, yet these values could represent a good reward function. Indeed, despite the reward being quite noisy for Seaquest once the score was removed, Table~\ref{atari_returns_tables} shows a policy trained on this noisy reward function obtains somewhat comparable evaluation performance on the environment's true reward as a policy trained in the environment with the score present. 

\begin{table}[ht]
  \caption{Return achieved by a PPO policy trained for 10M time steps, with the true environment reward vs.\ the output of a learned reward function. The learned reward functions were trained on observations with the score displayed.}
  \label{atari_returns_tables}
  \centering
  \begin{tabular}{lllll}
    \toprule
    Environment     & Return - trained w/ true reward     & Return - trained w/ regressed reward \\
    \midrule
    Breakout w/ Score Displayed & 360.3  & 141.9     \\
    Breakout w/o Score Displayed & 258.5  & 90.1     \\
    Seaquest w/ Score Displayed & 1746.0  & \textbf{712.0}     \\
    Seaquest w/o Score Displayed & 934.0  & \textbf{540.0}     \\
    \bottomrule
  \end{tabular}
\end{table}

\subsection{Summary of results}

We have demonstrated that our adapted saliency map techniques can identify features of the environment that influence reward function outputs.
This is sufficient to narrow down the space of hypotheses regarding how the reward function works.
These hypotheses can then be tested by counterfactual inputs, whether hand-crafted or via environment modification.
However, these methods don't directly predict how such changes effect the optimal policy(s) induced by the reward function.

Indeed, we saw in Figure~\ref{fig:maze_training_curves} and Table~\ref{atari_returns_tables} that even when the learned reward makes very different numerical predictions to the ground-truth, it may still incentivize similar policy behavior.
For example, a learned \texttt{CoinFlipGoal} reward function induces a good policy in \texttt{TwoGoals}, despite its values shifting significantly.
A learned reward for Seaquest can train a better-than-random policy even when its primary feature, the score, is removed making the reward function output much noisier.

The lack of predictive power for \emph{policy} behavior is a consequence of the fact that existing neural network interpretability techniques treat all changes in output as equally important.
However, some large changes (such as multiplying all outputs by a large positive constant) have \emph{no} effect on policy behavior.
Whereas some small changes (such as swapping the reward of the first and second-best states) will substantially change the policy.
Consequently, even the removal of some highly salient features from an environment, leading to totally different predicted reward values, can still leave the reward function capable of training a good policy in the modified environment. 

This limitation suggests that new methods are needed that incorporate a notion of difference taking into account the unique mathematical structure of reward functions.
Measures like EPIC, discussed in section~\ref{sec:evaluating_functions}, could be a useful starting point for this.

\section{Conclusions and Future Work}

In this paper, we have explored the use of saliency maps and counterfactuals to understand learned reward functions. We discovered that: (1) Sometimes reward functions rely on surprising information, and implement counterintuitive algorithms, in computing reward. (2) Basic insights into how learned reward functions work, such as the features they depend on, can be gained with existing interpretability techniques. However, (3) these techniques are poorly suited for predicting whether a reward function will transfer to new environments. While they can predict whether a reward function's output will change when aspects of the environment are changed, it is possible for the output to change but for the policy trained on the reward function in the new environment to still achieve high return on the new environment's true reward. In this sense, they are a conservative method: they will highlight major differences, but also many irrelevant ones.

Taken together, these results underscore the need for a new line of work: techniques for verifying whether learned reward functions truly reflect a user's preferences. Such work appears both tractable and important. It is tractable because the complexity of our objectives is likely to remain fixed, while the capability of ML systems (including RL agents) will grow over time. But it is also important -- as we seek to deploy advanced RL systems in increasingly complex real-world environments, we will rely on reward learning algorithms to save us from having to manually specify an objective. To ensure that this can be done safely, we need ways of verifying that these learned reward functions truly reflect our objectives -- in the words of Norbert Wiener, that they reflect "the purpose which we really desire and not merely a colorful imitation of it"~\cite{wiener1960some}.

\section*{Broader Impact}
When training RL agents for complex real-world tasks, it is important that the reward function the agent is trained on truly represent human preferences. But because it is difficult to manually translate the totality of human preferences into a reward function, reward learning should be used instead -- human preferences should be \emph{learned}. In this paper, we explored some techniques for identifying when reward learning has failed. We hope that these methods, and ones like them, will be useful in auditing learned reward functions for real-world tasks. 

A potential downside we see from our work is that our interpretability methods could give one a false sense of security. While our methods \emph{can} elucidate problems, they are \emph{not guaranteed} to reveal problems in learned reward functions. Much further work is needed, and engineers should not place too much trust in the techniques explored in this paper.

\begin{ack}
We thank Cody Wild for her feedback on previous drafts. We thank researchers at the Center for Human-Compatible AI for helpful discussions. Work supported in part by the Berkeley Existential Risk Initiative.
\end{ack}

\bibliography{refs.bib}


\end{document}